\documentclass[graybox, envcountchap, natbib]{svmult}

\usepackage{hyperref}
\usepackage{mathptmx}        
\usepackage{helvet}          
\usepackage{courier}         
\usepackage{makeidx}         
\usepackage{graphicx}        
\usepackage{multicol}        
\usepackage[bottom]{footmisc}
\usepackage[utf8]{inputenc}
\usepackage{mathtools}
\usepackage{tcolorbox}
\usepackage{rotating}

\usepackage[nonumberlist,nopostdot,xindy,toc,acronym]{glossaries}
\setglossarystyle{altlist}

\newglossaryentry{coopetition}{
    name={coopetition},
    description={TODO}
}

\newglossaryentry{classification}{
    name={classification},
    description={A \gls{supervised-learning} task  in which labels are categorical variables (discrete, usually non ordinal).}
}

\newglossaryentry{regression}{
    name={regression},
    description={A \gls{supervised-learning} task in which labels are continuous or ordinal (categorical regression).}
}

\newglossaryentry{dataset}{%
  name={dataset},%
  description={A collection of labeled or unlabeled \gls{data-samples}.}%
  }

\newglossaryentry{dataset-requirements}{
    name={dataset requirements},
    description={The set of specifications of a dataset outlined in a document drafted before and collection begins. It should include, among other things, nature of the data, purpose, quality and quantity, means, costs, timeline.}
}

\newglossaryentry{dataset-design}{%
  name={dataset design},%
  description={The formal definition of a particular \gls{data-collection}.}.
  }
  
\newglossaryentry{dataset-implementation}{%
  name={dataset implementation},%
  description={Carrying out a particular \gls{data-collection} according to a given \gls{dataset-design}.}.
  }
  
\newglossaryentry{dataset-life-cycle}{
    name={dataset life-cycle},
    description={The set of processes which creates or transforms a dataset from \gls{dataset-requirements} to its \gls{dataset-distribution} and \gls{dataset-maintenance} or \gls{dataset-deprecation}.}
}

\newglossaryentry{data-collection}{
    name={data collection},
    description={The set of processes involved in producing a \gls{dataset}.}
}

\newglossaryentry{data-gathering}{
    name={data gathering},
    description={A particular \gls{data-collection}, which consists in gathering existing \gls{data-samples} or an entire existing \gls{dataset}, possibly from different sources, on which we do not have any influence.}
}

\newglossaryentry{data-reusing}{
    name={data reusing},
    description={The action of using an existing \gls{dataset} in an application, without changing its original purpose. Reusing is distinct from \gls{data-repurposing} or \gls{data-recycling}.}
}

\newglossaryentry{data-repurposing}{
    name={data repurposing},
    description={The action of using an existing \gls{dataset} in a new task or application, by changing its original purpose. Repurposing may include changing \gls{data-annotation}, re-sampling, transforming. Repurposing is distinct from \gls{data-reusing} or \gls{data-recycling}.}
}

\newglossaryentry{data-recycling}{
    name={data recycling },
    description={A particular \gls{data-collection}, which consists in recycling existing \gls{data-samples} or an entire existing \gls{dataset}, possibly from different sources, on which we do not have any influence. Recycling is distinct from \gls{data-reusing} or \gls{data-repurposing}.}
}

\newglossaryentry{data-resampling}{
    name={data resampling},
    description={The action selecting particular \gls{data-samples} in a \gls{dataset}. This may yield \gls{sample-bias} (or correct it).}
}

\newglossaryentry{synthetic-data-generation}{
    name={synthetic data generation},
    description={Is the process of generating artificial data that mimics real-world observations and can be used to (pre)train machine learning models when actual data is difficult or expensive to get.}
}

\newglossaryentry{data-bias}{
    name={data bias},
    description={The existence of spurious correlations (or dependencies) between data labels \gls{spurious-feature}s. Data bias includes \gls{sample-bias} and \gls{confounding-bias}}
}

\newglossaryentry{sample-bias}{
    name={sample bias},
    description={A particular type of \gls{data-bias} induced by selecting samples in a non-representative way of the target population.}
}

\newglossaryentry{confounding-bias}{
    name={confounding bias},
    description={A particular type of \gls{data-bias} induced by the existence of a common cause between labels and \gls{spurious-feature}s.}
}

\newglossaryentry{spurious-feature}{
    name={spurious feature},
    description={A variable (or feature or attribute) of \gls{data-samples}, which may be a ``protected attribute'' that could trigger discrimination (\eg gender, ethnicity, or age) or a variable known to be irrelevant to the particular task at hand (\eg the identity of the data collector, the time of data collection, the recording conditions).}
}

\newglossaryentry{core-feature}{
    name={core feature},
    description={A variable (or feature or attribute), which is though of as ``leggitimate'' to use to perform a \gls{learning-task}, as opposed to a \gls{spurious-feature}. Note that it is possible, because of \gls{entanglement} that some variable in a \gls{data-represention} combine original spurious and core features. Merely suppressing spurious features may not resolve the problem of bias in data.}
}

\newglossaryentry{data-simulation}{
    name={data simulation},
    description={A particular \gls{data-collection}, which consists in artificially creating data with a numerical simulator.}
}

\newglossaryentry{data-acquisition}{
    name={data acquisition},
    description={A particular \gls{data-collection}, which consists in measuring a real world signal with instruments, and converting it to digital information.}
}

\newglossaryentry{data-annotation}{
    name={data annotation},
    description={The action of attaching to existing data another type of complementary data, often called \gls{meta-data}, providing extra information (\eg object bounding boxes).}
}

\newglossaryentry{meta-data}{
    name={meta-data},
    description={Various \gls{data-annotation}s providing extra information to \gls{data-samples} (\eg object bounding boxes, date and time of collection, data origin).}
}

\newglossaryentry{data-labelling}{
    name={data labelling},
    description={A particular \gls{data-annotation}, which consists in mapping data samples to class labels. By extension, in the case of regression, labels can be continuous variables.}
}

\newglossaryentry{data-samples}{
    name={data samples},
    description={Individual units in a dataset.}
}

\newglossaryentry{data-transformation}{
    name={data transformation},
    description={A set of processes, which map an input data representation to an output data representation.}
}

\newglossaryentry{data-augmentation}{
    name={data augmentation},
    description={A set of processes which generates new data samples based on a fixed input dataset.}
}

\newglossaryentry{data-integration}{
    name={data integration},
    description={A set of processes which merge different datasets with complementary properties such as features, sources or quantities.}
}

\newglossaryentry{data-fusion}{
    name={data fusion},
    description={Synonym of data-integration.}
}

\newglossaryentry{data-cleaning}{
    name={data cleaning},
    description={A set of processes which detect, remove or replace mistakes in data samples.}
}

\newglossaryentry{data-reduction}{
    name={data reduction},
    description={A set of processes which reduce the quantity of information contained in a dataset for example to have improve computational efficiency or to reduce noise.}
}

\newglossaryentry{data-representation}{
    name={data representation},
    description={A well specified data structure in which each \gls{data-samples} is stored in a digital form in a computer. Feature vectors are a frequently used data representation in Machine Learning.}
}

\newglossaryentry{data-normalization}{
    name={data normalization},
    description={...}
}

\newglossaryentry{data-calibration}{
    name={data calibration},
    description={...}
}

\newglossaryentry{dataset-evaluation}{
    name={dataset evaluation},
    description={A set of processes, which verify that the produced \gls{dataset} meets the \gls{dataset-requirements}.}
}

\newglossaryentry{dataset-distribution}{
    name={dataset distribution},
    description={Means of making a \gls{dataset} available to the public.}
}

\newglossaryentry{dataset-maintenance}{
    name={dataset maintenance},
    description={Means of ensuring error corrections, upgrades, user-feedback, new regulations, or new ethical standards, are incorporated into new versions of the dataset.}
}

\newglossaryentry{dataset-retirement}{
    name={dataset retirement},
    description={Retiring a \gls{dataset} from public availability because of irreversible errors, new regulations, or new ethical standards.}
}

\newglossaryentry{dataset-deprecation}{
    name={dataset deprecation},
    description={Notification that a \gls{dataset} should not be used anymore but remains public for traceability or restricted usage because of irreversible errors, new regulations, or new ethical standards.}
}

\newglossaryentry{data-soundness}{
    name={data soundness},
    description={...}
}

\newglossaryentry{data-completeness}{
    name={data completeness},
    description={...}
}

\newglossaryentry{data-compactness}{
    name={data compactness},
    description={...}
}

\newglossaryentry{learning-task}{%
  name={learning task},%
  description={A formal prediction problem, denoted as $T$, composed of \gls{training-data} and \gls{validation-data} or \gls{test-data}, depending on the challenge phase, with the objective of (1) producing a \gls{predictor} using a \gls{trainer} (supplied by a challenge participant) using training data, and (2) evaluating the \gls{predictor}, using validation or test data. The evaluation is carried out using one (or several) \gls{evaluation-metric}s.}%
}

\newglossaryentry{evaluation-metric}{%
  name={evaluation metric},%
  description={A function $L(\hat{y}, y)$, of a prediction  $\hat{y}=f(x)$ made by a \gls{predictor} $f(.)$ on a input sample $x$ (from \gls{validation-data} or \gls{test-data}) and an associated ``ground truth'' value $y$. A score is obtained by averaging $L(\hat{y}, y)$ over all validation of test samples, which estimates empirically the generalization performance of the predictor. }
  }

\newglossaryentry{training-data}{%
  name={training data},%
  description={A subset of the \gls{dataset} $D_{tr} = \{ (x_i, y_i) \}_{i=1}^n \subseteq X \times Y$, which represent a snapshot of the ``real-world'', assumed to be drawn from a probability distribution $P$ on $X \times Y$, which represents the abstract process generating the data (usually unknown). This subset is used by the \gls{trainer} (or learning algorithm).}%
}

\newglossaryentry{validation-data}{%
  name={validation data},%
  description={A subset of the \gls{dataset}, disjoint from the \gls{training-data}, but usually drawn from the same distribution, used to get feed-back on performance during the development of a learning algorithm, so the \gls{test-data} remains untouched until the final evaluation. Using test data to adjust hyper-parameters of select algorithms results in optimistically biased performances. In a challenge, validation data are used for evaluation during the \gls{development-phase} (or feed-back phase).}%
}

\newglossaryentry{warmup-phase}{%
  name={warmup phase},%
  description={A challenge phase (also called public phase or challenge beta-testing) during which the participants are invited to try out the challenge protocol before the official starting date of the \gls{development-phase}, using some sample (public) data, not used in subsequent phases.}
  }

\newglossaryentry{development-phase}{%
  name={development phase},%
  description={A challenge phase (also called feed-back phase), during which the participants develop a solution to the problem of the challenge and submit results for immediate evaluation on \gls{validation-data}, to received feed-back on a (public) leaderboard.}
  }
  
  \newglossaryentry{final-phase}{%
  name={final phase},%
  description={A challenge phase (also called final test phase), during which the participants are evaluated on \gls{test-data}, not used either for training or validation. The final phase leaderboard remains private (only visible to the organizers) until the challenge is over.}
  }
  
\newglossaryentry{test-data}{%
  name={test data},%
  description={A subset of the \gls{dataset}, disjoint from the \gls{training-data} and the \gls{validation-data}, used to get feed-back on performance during the development of a learning algorithm, so the \gls{test-data} remains untouched until the final evaluation. Using test data to adjust hyper-parameters of select algorithms results in optimistically biased performances. In a challenge, test data are used for evaluation only once (for each final entry of each participant) in the \gls{final-phase}.}%
}

\newglossaryentry{alpha-challenge}{%
  name={$\alpha$-challenge},%
  description={}%
}

\newglossaryentry{beta-challenge}{%
  name={$\beta$-challenge},%
  description={}%
}

\newglossaryentry{predictor}{%
  name={predictor},%
  description={A function $f(.)$ mapping an input $x$ to an output $y$, solving a {classification} or a \gls{regression} problem. An \gls{alpha-challenge} asks participants to submit (trained) predictors.}%
}

\newglossaryentry{trainer}{%
  name={trainer},%
  description={An algorithm (usually called learning algorithm), which outputs a \gls{predictor}, or more generally a trained learning machine, which may be a data generator. The input to a trainer is \gls{training-data}. A \gls{beta-challenge} asks participants to submit trainers, and eventually tests on multiple \gls{learning-task}s their capability of automated machine learning (AutoML).}%
}

\newglossaryentry{meta-trainer}{%
  name={meta trainer},%
  description={An algorithm, which outputs a \gls{trainer}, given a dataset of datasets (meta-training set). A \gls{gamma-challenge} asks participants to submit meta-trainers, and tests on a meta-test set their capability of learning to learn (meta-learning).}%
}

\newglossaryentry{supervised-learning}{
    name={supervised learning},
    description={A \gls{learning-task} in which labels $Y$ are specified.}
}

\newglossaryentry{power-differential}{
    name={power differential},
    description={The difference in power between persons in positions of authority and those individuals in subordinate positions, which results in a vulnerability on the part of the subordinate. For example, the natural differences in power that exist between faculty and student.}
}

\newglossaryentry{observational-setting}{
    name={observational setting},
    description={A method for collecting data in which the investigator in charge of data collection does not interfere with the phenomenon. The distribution of samples collected should reflect the ``natural'' distribution of data.}
}

\newglossaryentry{experimental-setting}{
    name={experimental setting},
    description={A method for collecting data in which  the investigator  interferes with the natural world to achieve desired effects. A planned experiment consists in varying some factors systematically, in a controlled manner, or randomly.}
}

\newacronym{dlc}{DLC}{Data Life-Cycle}

\newacronym{ddlc}{DDLC}{Dataset Development Life-Cycle}

\newacronym{gdpr}{GDPR}{General Data-Protection Regulation}

\newacronym{mllc}{MLLC}{Machine Learning Life-Cycle}

\newacronym{sari}{SARI}{State Action Reward Information}

\newacronym{dream}{DREAM}{Dialogue for Reverse Engineering Assessment and Methods}

\newacronym{casp}{CASP}{Critical Assessment of protein Structure Prediction}

\newacronym{miccai}{MICCAI}{Medical Image Computing and Computer Assisted Intervention Society}
\makeglossaries

\usepackage{xspace}
\newcommand*{\eg}{e.g.\@\xspace}

\usepackage{pdfpages}
\usepackage{amssymb,graphicx,amsmath} 
\usepackage{times}
\usepackage{tikz}
\usepackage{soul}
\usepackage{color}
\usepackage{xcolor}

\usepackage{enumerate}
\usepackage{comment}

\definecolor{MyDarkGreen}{rgb}{0.17,0.46,0.25} 
\definecolor{MyDarkRed}{rgb}{0.88,0.22,0.21} 
\definecolor{MyDarkBlue}{rgb}{0.11,0.11,0.70} 
\definecolor{lightgray}{gray}{0.85}
\definecolor{blue}{rgb}{0.1216, 0.4667, 0.7059}
\definecolor{orange}{rgb}{1.0, 0.4980, 0.0549}
\sethlcolor{Dandelion}

\usetikzlibrary{arrows,shapes,plotmarks,positioning}
\usetikzlibrary{decorations.markings}

\tikzset{>=stealth'} 
\tikzstyle{graphnode} = 
   [circle,draw=black,minimum size=22pt,text centered,text
     width=22pt,inner sep=0pt] 
\tikzstyle{var}   =[graphnode,fill=white]
\tikzstyle{vardashed}   =[graphnode,draw=gray,fill=white]
\tikzstyle{obs}   =[graphnode,fill=black,text=white]
\tikzstyle{obsgrey}   =[graphnode,draw=white,fill=lightgray,text=black]
\tikzstyle{par}    =[graphnode,draw=white,fill=red,text=black] 
 \tikzstyle{crucial} =[graphnode,draw=white,fill=yellow,text=black] 
\tikzstyle{fac}   =[rectangle,draw=black,fill=black!25,minimum size=5pt]
\tikzstyle{facprior} =[rectangle,draw=black,fill=black,text=white,minimum size=5pt]
\tikzstyle{edge}  =[draw=white,double=black,very thick,-]
\tikzstyle{blueedge}  =[draw=white,double=blue,very thick,-]
\tikzstyle{rededge}  =[draw=white,double=red,very thick,-]
\tikzstyle{prior} =[rectangle, draw=black, fill=black, minimum size=
5pt, inner sep=0pt]
\tikzstyle{dirprior} = [circle, draw=black, fill=black, minimum
size=5pt, inner sep=0pt]

\usepackage{graphicx,subcaption} 
\usepackage{tikz}
\usetikzlibrary{calc,arrows}
\usepackage{color, colortbl}
\usepackage{booktabs} 
\let\proof\relax

\usepackage{amsthm}
\usepackage{amssymb}
\usepackage{array} 
\usepackage{paralist} 
\usepackage{verbatim} 
\usepackage{makecell}
\usepackage{array,multirow}
\usepackage{mathrsfs}
\usepackage[T1]{fontenc}
\usepackage{varwidth}
\usepackage{hyperref}
\usepackage{epigraph}

\usepackage{here}
\usepackage{dsfont}
\usepackage{amsfonts}
\usepackage{pifont}

\usepackage{forest}

\usepackage[disable]{todonotes}
\newcommand{\adrien}[1]{\todo[inline,color=green!40]{#1 -- Adrien}}
\newcommand{\isabelle}[1]{\todo[inline,color=orange!40]{#1 -- Isabelle}}

\usepackage{eurosym}
\usepackage[graphicx]{realboxes}
\usepackage{pdflscape}

\usepackage[lined, ruled, boxed, linesnumbered, commentsnumbered]{algorithm2e}

\title{Practical issues: Proposals, grant money, sponsors, prizes, dissemination, publicity}
\author{Magali Richard, Yuna Blum, Justin Guinney,  Gustavo Stolovitzky, and Adrien Pav\~ao}
\authorrunning{Magali Richard et al.} 
\institute{TIMC UMR 5525, Univ. Grenoble Alpes, CNRS \email{magali.richard@univ-grenoble-alpes.fr}}
\begin{document}

\maketitle

\label{chap:practical}

\abstract*{Abstract}
\abstract{Each organization of competitions and benchmarks involves a large number of practical problems, such as obtaining sufficient financial support or recruiting participants through appropriate incentives and community engagement. In addition to defining scientific tasks, preparing data and creating challenges, a very important practical administrative organization remains to be achieved. Indeed, cost assessment, corresponding requests for financial support and adequate publicity are key factors for successful organization of the competition. In addition, a good understanding of the incentives that lead participants to engage in a given challenge is fundamental for effective practical organization success. In this chapter, we will cover these topics and give some practical tips and examples for overcoming the ``challenge'' of organizing the challenges.}

\keywords{communication, proposal, sponsors, practical issues}

\isabelle{This chapter is nice and cover lots of relevant topics. It is missing references and pointers to practical resources.}

\newpage

\section{Incentivizing participation}
\label{ch12:sec:incentivizing}

\begin{figure*}[t]
\centering
\includegraphics[width=\linewidth]{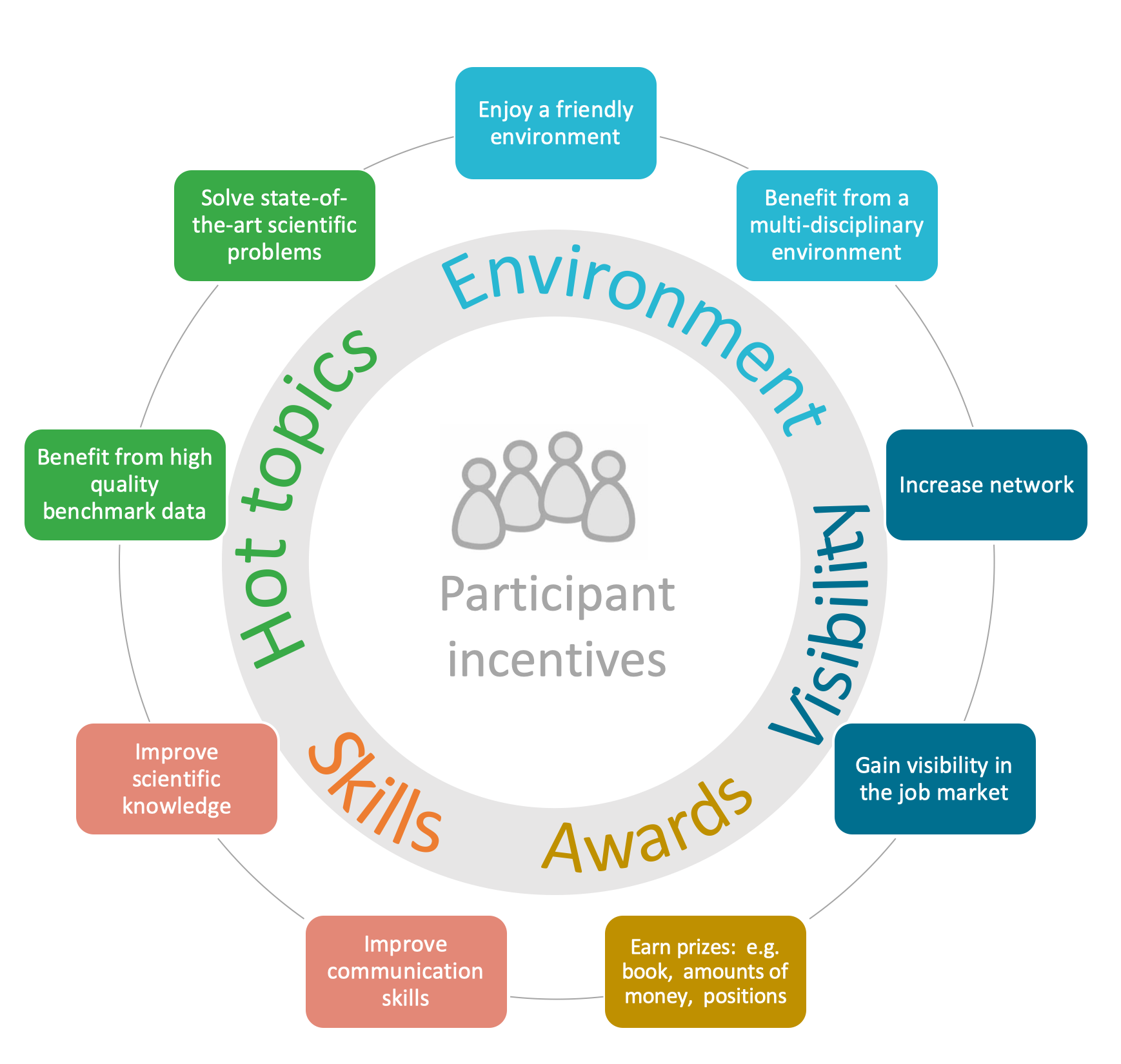}
\caption{The incentives for participating in a challenge.}

\label{figure:1}
\end{figure*}

How to incentivize participants to work on complex problems is a key feature of challenge organization. In this section, we review various types of motivations (Figure \ref{figure:1}), from a participant perspective, and give practical tips to optimize those incentives. 

\subsection{Skills : Knowledge acquisition, communication, education}

Traditional university programs in Artificial Intelligence are evolving rapidly, trying to meet the new needs of students, especially on their ability to work collaboratively while improving their scientific knowledge on data mining. Data challenges are mainly based on a coopetitive model, which has the advantage of responding to this dual motivation. Coopetition \cite{brandenburger_co-opetition_2011} is an active learning pedagogical approach based on the combination of a strategy of competition, where students compete for the best result, and cooperation, where students collaborate for a mutual benefit. Coopetition-based data challenges have the advantage of simultaneously offering two types of learning. On the one hand, this gives a participant a solid methodological training on the scientific question addressed, thanks to the sharing of knowledge between professors and students, but also between the students themselves. On the other hand, these approaches allow students to acquire new skills in collaboration, communication and networking. For more details, please refer to chapter 9: Competitions and challenges in education.

Educational data challenges can be organized into teamwork, recruiting participants from different backgrounds (academic and cultural), with a scientific preparation that can range from minimal information about the challenge before starting to full preparation through a series of dedicated conferences. To meet the expectations of the students, a key factor is the will of the organizer to build a "friendly environment" which will help to boost the motivation of the students and their self-esteem, and to focus more on the process itself than on the results and objectives.
Building multidisciplinary teams with different scientific expertise and focusing on real problems are important aspects in the organization of educational challenges. It is also important to provide an environment where participants can communicate with their team members, other teams, and teachers. Setting the right reward and price is a major motivator for winning student buy-in \cite{abernathy_academic_2001}. 

Finally, organization of competitions itself can be used as a pedagogical tool. Designing such task is complex and can be, in some regards, more interesting than solving it \cite{pavao2019design}.

\subsection{Hot topics : scientific crowdsourced benchmarking}

The quintessential challenge revolves around an existing quantitative standard or benchmark, and seeks to improve upon state-of-the-art. One of the more longstanding benchmark initiatives is the Critical Assessment for Structural Proteins (CASP), which asks participants to predict protein structure (folding) from protein sequence. Groups who specialize in this domain are naturally incentivized to compare their approach in the structured and objective format of a data challenge in the hope that their method out-competes other approaches and can therefore become a new standard in the field \cite{bender_challenges_2016}. CASP is now recognized within the protein structure community as the \textit{de facto} forum for assessing algorithms, and is therefore as much an incentive as a mandate for formal recognition with the community. This incentive generalizes to all specialties, including image recognition (e.g. MNIST \cite{madry_towards_2019}, ImageNet \cite{russakovsky_imagenet_2015}), gene identification and function prediction (e.g. RGASP \cite{steijger_assessment_2013}, CAFA \cite{radivojac_large-scale_2013}) or drug binding (e.g. \href{https://www.hiit.fi/idg-dream-challenge/}{on going DREAM drug binding challenge}). 

Any published AI algorithm is expected to include a formal performance comparison against state-of-the-art methods. No good data-driven approach could emerge without good quality, well curated data. This task can be cumbersome and require a great deal of work to assemble and prepare benchmark datasets. Depending on the type of data, data acquisition and/or generation can be very time-consuming and costly (see cost section below).  Consequently, a natural perk of a scientific data challenge is that the work involved to generate and prepare a benchmarking dataset is managed by the challenge organizers. Therefore, AI competitions offer a playground with data that are usually costly and complicated to generate. Access to these types of datasets is a strong motivation for participants aiming to develop cutting edge methodological approaches to solve a complex scientific problem. 

Recurrent challenges also present the advantage or keeping people on a regular schedule, as they expect the challenge to come and reserve time for it. As foor a classic scientific event, it provides participants the opportunity to expand their professional network and to start new collaborations with people working in the same research field or people from different disciplines gravitating around the same topic. Finally, data challenges remain the best functioning way of implementing coopetitions: people compete and get credit for winning, then they share their solution publicly and the community can move together to the next step.

\subsection{Environment and awards}

One appealing aspect of the challenges is the spirit of games. This translates into a friendly yet competitive environment along with rewards. It is not unusual to gather common participants on different challenges. An addiction for this type of competition can emerged as well as a thrill of seeing evolution in the social community (particularly true for commercial platforms such as Kaggle). The rewards can be of various nature going from small prizes (e.g. book) to high amounts of money (e.g. 1 million dollars, Salesforce 1 Hackathon) or even positions in companies. Big awards may naturally attract more participants but this has to be balanced against the environment of the challenge and the scientific problem raised.

\subsection {Visibility, career and recruitment}

Challenges are opportunities for participants to showcase their various skills to recruiters and even get a position at the end. A growing number of organizations are adopting modern hiring practices such as challenges to find best candidates. Recruiters use this tool to assess candidates' technical and behavior skills. Challenges have indeed the great advantage of evaluating many different criteria at the same time.  Companies can assess technical competencies such as problem solving skills, time management and innovations. They can also assess the behavioral skills they value, such as communication, openness to diversity and leadership.  

The implementation of a challenge allows recruiters to define certain expectations towards the evaluated candidates (candidates gain insight into the work culture of their future employer), while verifying if their personality corresponds to the company's fundamental values. One of the difficulties in recruitment is that many companies still follow long selection processes that waste time and interest for both candidates and recruiters. To overcome this problem, challenges can be used to evaluate candidates in a short period of time and friendly environment, where they can demonstrate real-time expertise. It can also serve as a pre-selection process that will also save time for recruiters. As a result, such challenges must be carrefully designed according to the skills (technical, innovation, leadership ...) one wants to evaluate.

Interestingly, challenges can bring together a larger number of candidates from more diverse backgrounds than traditional recruiting. Organizers can build a portfolio of interesting candidates for present and future positions, without necessarily limiting themselves to the winners of the challenge, especially as winning challenges does not include all skills valuable in project management (such as communication with customers or production standards). For instance, Kaggle,one of the leading challenges platform acting as a recruiting tool, usis a performance tracking system to evaluate participants\footnote{\url{https://www.kaggle.com/progression}}. Some companies even sell expertise from Kaggle Grandmasters\footnote{\url{https://h2o.ai/company/team/kaggle-grandmasters/}}. Besides, challenges are also an excellent way to increase brand awareness. They can be used as a marketing tactic for big companies to reinforce their leadership in their field. Smaller companies can also increase their visibility though challenges and attract more applicants for a position.

Finally, in addition to recruiting new talent, challenges allow companies to bring innovative solutions and ideas to technical problems.
Based on the clear success of challenges in the recruitment process, we can easily expect their increase in the upcoming years.

\begin{tcolorbox}
\textbf{Practical tips and resources to optimize incentivization}
\begin{itemize}
   \item Define your working plan and your objectives\footnote{10 tips here: http://www.chalearn.org/tips.html}
  \item Carefully prepare benchmarking datasets (see Chapter 3 on data preparation).
   \item Set up a website to collect a list of interested people\footnote{see e.g. https://l2rpn.chalearn.org/}.
   \item Bring together a expert steering committee
   \item Provide good educational materiel together with the challenge (i.e.  a good starting kit, white paper). 
   \item Make yourself available during the challenge to answer questions. 
   \item Be responsive to questions on the forum.
   \item For a recurrent challenge,  provide open-source previous winning solutions.
   \item Organize good publication venues (see details and examples in section 12.2 Community engagement) 
  \item Associate with established conferences (see details and examples in section 12.2 Community engagement) 
   \item For education challenges, you can find inspiration on existing education challenges on open-source platforms such as RAMP\footnote{https://ramp.studio}, or Codalab\footnote{https://codalab.lisn.upsaclay.fr/}
\end{itemize}
\end{tcolorbox}

\section{Community engagement}

\begin{figure*}[t]
\centering
\includegraphics[width=\linewidth]{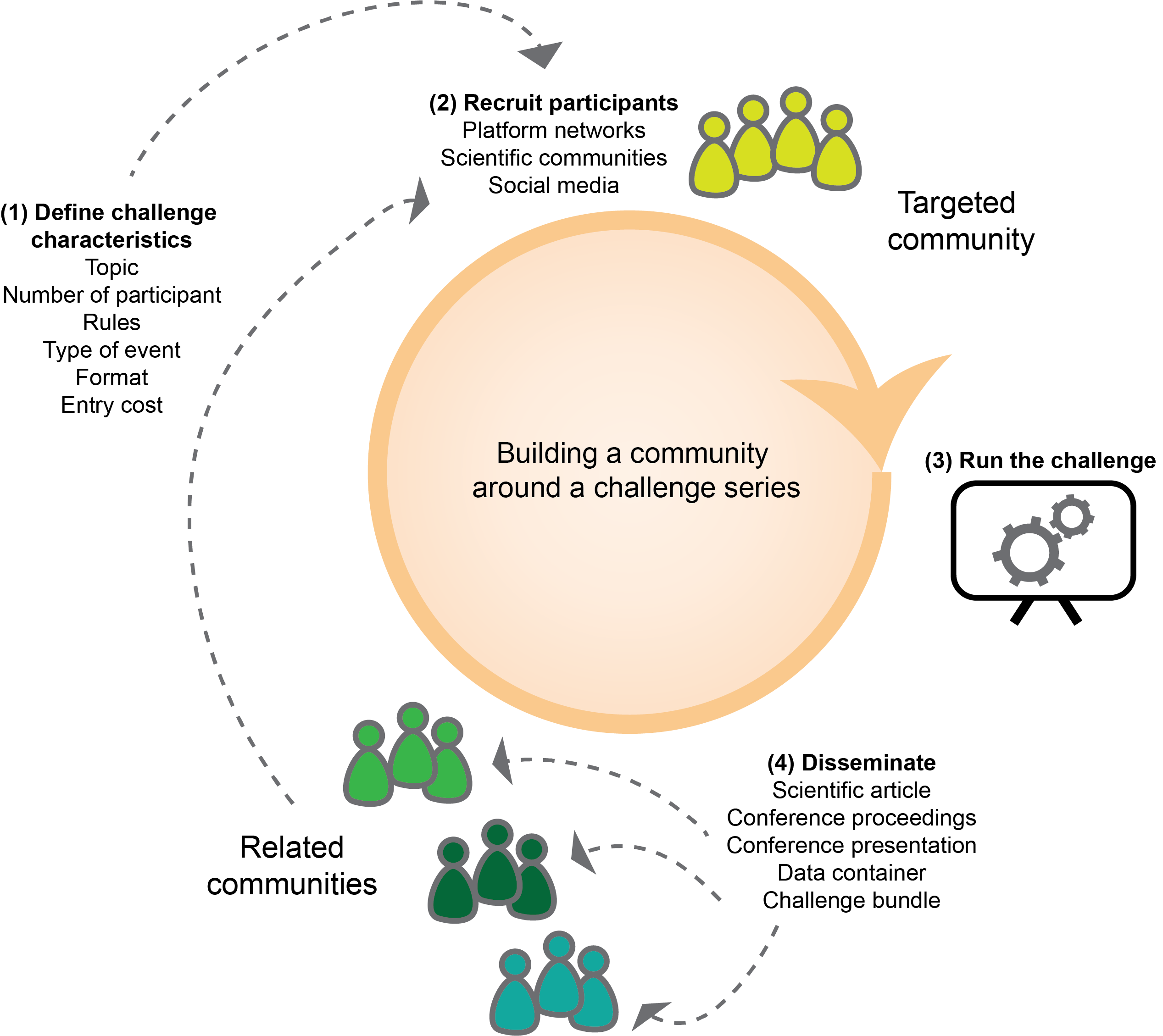}
\caption{The process of engaging a community}
\label{figure:2} 
\end{figure*}

Mechanisms for engaging and disseminating a competition towards a targeted community are complex and highly dependant on the scientific field. In this section, we try to review general aspects of community engagement that could help challenge organizers to properly define their strategy. See Figure \ref{figure:2}  and Table \ref{table:community} for a review of community engagement strategies and examples of recent competitions.

\subsection{Organization of the challenge}

The community that will engage in a specific competition will depend on several key aspects of defining the challenge. First, the organizers should define an optimal number of participants (which includes the maximum number of participants allowing the challenge to be held properly). Large open competitions have the advantage of ensuring visibility and optimizing scientific production (in the case of crowdsourced benchmarking for example) while smaller competitions will promote communication between participants (more adapted to challenges aiming at educational results). Interestingly, recent challenges run a pre-selection to avoid crowding the final run, which may be expensive in computational resources. Then they have to determine an entry cost: is it easy to participate in the competition? The entry cost depends on several factors: clarity of the rules, specificity of the tasks, size of the dataset, computational resources required to run the methods... All of this will have an impact on the participants who will enter the competition and indirectly define the target audience. Finally, the organizers should established what the format of the competition will be: online events will increase the chances of getting a large pool of participants while in-person events (e.g. at dedicated schools or at scientific conferences) are more suitable for collaborative team work. Once all these parameters are specified, organizers can adapt their communication strategy accordingly and start communicating through dedicated channels, such as the scientific communities mailing list, the digital challenge platform networks and the social media.

\subsection{Challenge output dissemination}

The dissemination of the data challenge can take several formats (complementary and not exhaustive) and should match the following question : how would it serve the targeted community? 

Participatory benchmarking competitions generally result in scientific publications (see examples \cite{creason_community_2021,hadaca_consortium_guidelines_2020,marbach_wisdom_2012, eicher_challenges_2019,marot_learning_2021,le_artificial_2019}) which will be of use to the community. Offering authorship to competing teams, along with participation in manuscript design and writing, is also a strong incentive that will provide international visibility and recognition to participants. Organizers might try to connect with high-profile journal editors ahead of the challenge organization to discuss the possibility  of publishing the competition outcome. Depending on the scientific field of the competition, publications can take various form, such as scientific articles, contributions to special issues, conference proceedings, or even books. Best performing teams can also be offered the ability to present their solution in a international scientific conference (e.g. RECOMB/ISCB for DREAM deconvolution challenge). In addition to an article describing the results of the competition, a challenge built on the data to modeler model \cite{guinney_alternative_2018} could also result in publishing the benchmark dataset along with a container providing a reproducible and continuous benchmark (e.g. a dedicated docker container).  Competition data can then be re-used by research scientists as gold standard for new computational methods that will be developed in the future. Challenge organizers may also consider giving open access to their challenge design and templates, especially regarding educational challenges, so that these competition can be massively disseminated to various universities at no cost. 

Challenge output and dissemination strategy differ a lot according to the competition organizers and environments. Academic competitions massively rely on the open science framework, encouraging participants to submit their code under an open source license (ex: L2RPN, DREAM challenges). On the opposite, private companies are often motivated by solving an theoretical and methodological obstacle in order to further develop private commercial solutions that will be put on the market. Such organizers may be more inclined to follow a 'private output' model where participant surrender intellectual property of their findings in exchange for earning money prizes.  

\newpage
\begin{landscape}
\begin{table}[h!]
\begin{centering}
\begin{tabular}{ |p{5cm}||p{1.5cm}||p{1.5cm}||p{1.5cm}||p{2cm}||p{4cm}| }
 \hline
 \multicolumn{6}{|c|}{COMMUNITY ENGAGEMENT} \\
 \hline
Name & Field & Year & Platform & Number of participants & Dissemination \\
 \hline
TrackML Particle Tracking Challenge  & Physics & 2018 & Kaggle    & 739 participants &  IEEE WCCI competition (Rio de Janeiro, Jul 2018) and NIPS competition (Montreal, Dec 2018) \\
LAP series  & Computer Vision & 2013-22 & CodaLab   & XXX & Springer Series on Challenges in Machine Learning, ECCV, IEEE TPAMI, JMLR, IJCV, PAA, CVPR \\
Tumor Deconvolution  & Health & 2019\-20 & DREAM  &  38 teams &  2019 RECOMB/ISCB Regulatory and Systems Genomics, BiorXiv \\
AutoDL series (6 competitions so far) & Automated Machine Learning & 2019-21 & CodaLab  & more than 300 teams & ECML/PKDD, ACML, NeurIPS, IJCNN, WAIC, IEEE TPAMI \\
Digital Mammography & Health & 2017 & DREAM   & more than 120 teams & RECOMB/ISCB Regulatory and Systems Genomics \\
L2RPN  & Energy & 2020 & CodaLab  & more than 300 participants & NeurIPS, ArXiv\\
Challenge AI for industry  & Aeronotic & 2021 & CodaLab   & XXX & XXX \\
 \hline
\end{tabular}
\caption{Table of communities engagement}
\label{table:community}
\end{centering}
As a complement, a non-exhaustive list of conferences that have call for competitions, or can offer workshops and/or proceedings, as well as journals that can welcome competition result publication :

- \textit{Conferences and workshops}:
ESANN,
ICMLA,
WCCI (IJCNN, CEC),
ECML/PKDD (Discovery challenges),
KDD (KDD cup),
CVPR,
ECCV,
ECML/PKDD,
ICPR,
ICDAR,
IEEE international conference on big data,
IEEE International Conference on Automatic Face and Gesture Recognition (FG),
ACM SIGIR Forum,
NeurIPS dataset and benchmark track,
NeurIPS competition track,
Workshops @ NeurIPS, ICML, AAAI, CVPR, ICCV,
Workshop on Semantic Evaluation,
etc.

- \textit{Book series}: CiML Springer series, etc.

- \textit{Journals and pre-prints}:
International Journal of Forecasting,
International Journal of Information Retrieval Research (IJIRR),
IEEE Journal of Biomedical and Health Informatics,
IEEE Access,
Machine Vision and Applications,
IEEE TPAMI,
Nature methods,
Nature communication,
Journal of the American Medical Informatics Association,
Journal of biomedical informatics,
ArXiv,
BiorXiv,
PMLR,
BMC bioinformatics,
etc.

\end{table}
\end{landscape}

\newpage

\section{Costs, man power and resources}

Depending on the model chosen by the organizer, various costs will be associated with a competition organization. Human resources will also have to be invested to guarantee the quality of the organized challenge. To mitigate the problem of financing a competition, diverse sponsors, private companies or academic institutions can be involved. In this section, we review the costs, the need in man power and the resources the resources that can be requested when organizing challenges.

\begin{tcolorbox}
\textbf{An example of challenges costs: the L2RPN challenge / NeurIPS 2020}
\begin{itemize}
  \item \textbf{Research field} : Energy and environment.
  \item \textbf{Challenge platform} : Codalab\footnote{\url{https://competitions.codalab.org/competitions/25426}}.
  \item \textbf{Duration of the challenge} : 4 months.
  \item \textbf{Number of participants} : 300.
  \item \textbf{Data generation, access and curation : costs and resources description} : 70,000 euros.
  \item \textbf{Challenge engineering : costs and resources description} : 120,000 euros.
  \item \textbf{Challenge design, scientific expertise : costs and resources description} : 170,000 euros.
  \item \textbf{Prices, travel, conference organization (approximate evaluation of costs)} : 30,000 euros. 
  \item \textbf{Challenge governance (cost evaluation of legal, ethics and data privacy costs)} : none. 
  \item \textbf{Dissemination} : RTE, Google Research, University College of London, EPRI, IQT Labs. Chalearn. 
  \item \textbf{Sponsors} : PMLR\footnote{\url{https://arxiv.org/abs/2103.03104}} \& ChaLearn\footnote{\url{https://l2rpn.chalearn.org/}}
\end{itemize}
\end{tcolorbox}



\subsection{On overview of the requirments and associated costs}

\begin{figure*}[h!]
\centering
\includegraphics[width=\linewidth]{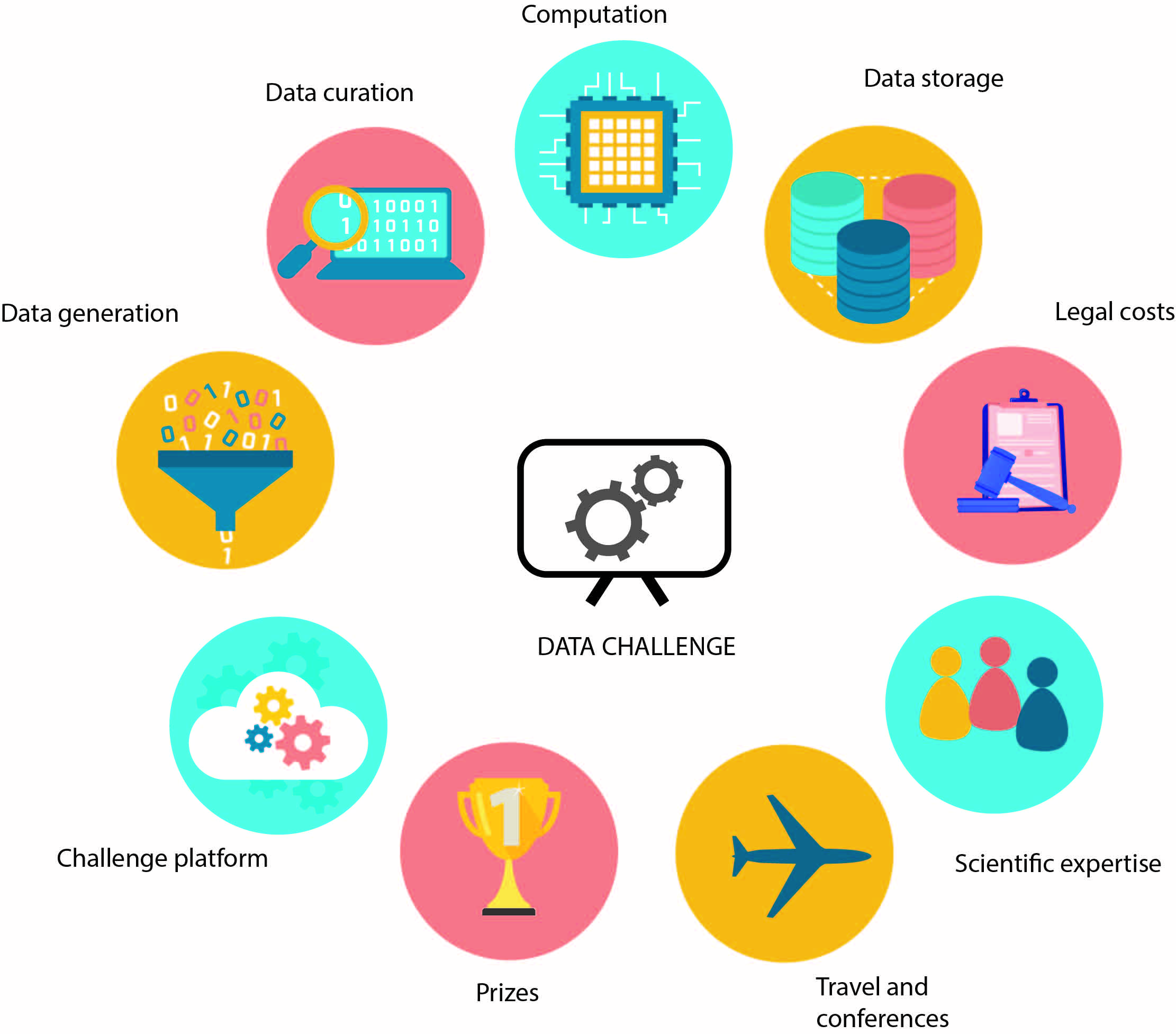}
\caption{Costs of data challenge organization. Pictures adapted from:  \href{https://fr.freepik.com/vecteurs/affaires}{macrovector}, \href{https://fr.freepik.com/vecteurs/or}{alvaro\_cabrera}, \href{https://fr.freepik.com/vecteurs/voyage}{visnezh} \&
\href{https://fr.freepik.com/vecteurs/abstrait}{vectorjuice} open work on freepik}
\label{figure:3}
\end{figure*}

\subsubsection*{Platform and registration system}

Several digital platforms can support challenge organization (see chapter 5 for different models of challenge platforms). Defining the platform should be a starting point of challenge organization, as open-source project such as CodaLab or commercial challenge platforms such as Kaggle will provide very different resources (technical support, engineering manpower dedicated to the competition...) and associated costs. Please refer to Chapter 5 for more details on the different services provided by each platform.

\subsubsection*{Data generation, access and curations}


Good quality, well curated data is a key factor of a competition success. General cost evaluation of data generation is complicated because it is highly variable depending on the scientific discipline involved. Data generation has always a cost, but this cost can bee supported by different players of the competition (sponsors, private companies, organisation committee, care providers, health insurance, etc). This costs also depends on the data type, size and accessibility. Data labelling can be achieved manually, using open source tools such as LabelMe, or private services such as Amazon Mechanical Turk in computer vision\footnote{A review of labelling tools can be found here: https://neptune.ai/blog/data-labeling-and-data-collection}
Good quality data also relies on the willing of organizers to work in synchronisation with the global efforts for technical standardization and ethic responsible data sharing, e.g. Global Alliance for Genomics and Health or FAIR principles for data management and stewardship \cite{wilkinson_fair_2016,cabili_simplifying_2018}. 

\subsubsection*{Governance and legal costs}

Competition governance strategy should also include legal counseling costs, that will ensure that the data storage and sharing concept complies with national and international legal requirements. In particular, usage of identifiable personal data (such as patient clinical data) is a complex and significant legal and data protection challenge \cite{nicol_consent_2019}. Moreover, rules for awarding prizes and travel grants should be clearly defined, this includes definitions of :

- jury's composition (committee of experts) 

- criteria of evaluation (e.g. relevance, usefulness, novelty, etc.)

- challenge submission process

- intellectual properties

- exclusion and appeal procedures

- control of the use of funds and goods, including prices

- privacy policy

- errors, frauds and breaches of rules mitigation plan




\subsubsection*{Computation and storage}

The digital data challenge platforms rely on cloud computing services to run and evaluate models. Access to these services can be externalized (such as Google Could Platform, Openstack, IBM Cloud or Amazon web services) or provided internally using the computing infrastructure of the challenge organizers. Depending on the competitions, the problem to solve and the type of data, the required resources vary a lot. For instance, in the case of code submission, it is important to estimate well the number of participants, and sometimes to limit the entries by setting a hard threshold. Indeed, code submission offers many advantages (controlled environments, confidential data, good sharing of the resources among participants, etc.) but is computationally very demanding. 
Thus, the organizers must accurately estimate the computation time of the expected methods as well as the type of computing units to use (\cite{ellrott_reproducible_2019}), knowing that donation of cloud units from Google, Azure and Amazon are relatively easy to obtain. Some platform such as Codalab can be coupled with such cloud services, via the use of compute workers. Finally, they need to decide accordingly whether they wish to offer computational services (allowing code submission) or ask participants to provide their own computational resources (only allowing the submission of results).

\subsubsection*{Scientific expertise, challenge design and engineering}

Bringing together an expert steering committee is a key factor to ensure that the issue raised by the competition corresponds to the needs of the community, and that the data will be used correctly to ask the right question. These two points are essential to ensure community engagement and the quality of the competition. In some specific occasion, building a realistic environment to simulate the different tasks of a competition can required a lot of work (research and engineer manpower ahead of the competition). For instance, L2RPN competition series required the generation of a dedicated framework and the generation of synthetic data with several people working on the project for over a year (cost of $\sim$200k\euro{}, see Table \ref{table:2}). Once the competition is completed, manpower is also needed to analyse the results, summarize, and disseminate the challenge outcomes.

\subsubsection*{Prizes, Travel and conference organization}

Reward costs should be included in the challenge budget. Prizes can be a important incentive to recruit participant (see section 1). . In case of in-person events, travel and conference organization costs should be considered. This can include speakers invitations, participation to the venue costs and travel grants for students. Example of costs to organized an event can be found in Table \ref{table:workshop}).

\begin{table}[H]
\begin{center}
\begin{tabular}{ |p{1cm} | p{7cm} | p{3cm}|}
 \hline
& Expense type & Estimated cost (EUR)\\
 \hline\hline
1 & Invited speakers registration (4x\$250) & 1,000 \\
 \hline
2 &Organizer travel expenses (3x\$2000) & 6,000 \\
 \hline
3 & Lunch (catering) for 40*\$50) & 2,00 \\
 \hline
 4 & Dinner for invited speakers/winners/organizers 20*\$50 & 1,000 \\
 \hline
\end{tabular}
\caption{Conference or workshop organization for a total budget of 10,000 euros.}
\label{table:workshop}
\end{center}
\end{table}

\subsection{Man power}

Man power is crucial in competition organization, and should not be underestimated. See Table \ref{table:manpower} for an average estimation of man power required to organize a challenge.
\begin{table}[]H]
\begin{center}
\begin{tabular}{ |p{1.5cm} | p{8cm} | p{1.5cm}|}
 \hline
Task& Description & Hours\\
 \hline\hline
1 & \textbf{Finding/reviewing data.} & 50 \\
 \hline
2 & \textbf{Formatting data.} Preprocess and format the data to simplify the task of participants, obfuscate the origin, anonymize. & 100 \\
 \hline
3 & \textbf{Assessment.} Define a task and evaluation metrics. Define and implement methods of scoring the results and comparing them. & 50 \\
 \hline
 4 & \textbf{Baseline software; starting kit.} Implement a simple example performing the tasks of the challenge. Prepare useful software libraries, make examples. & 100 \\
 \hline
 5 & \textbf{Result formats and software interfaces.} Define the formats in which the results should be returned by the systems and how experimentation will be conducted during the challenge. & 50 \\
 \hline
 6 & \textbf{Benchmark protocol.} Define the rules of the competition and determine the sequence of events. & 50 \\
 \hline
 7 & \textbf{Web portal.} Implement on challenge platform the benchmark protocol allowing on-line submissions and displaying results on a leaderboard.
 & 25 \\
 \hline
 8 & \textbf{Guidelines to participants.} Write the competition rules, document the formats and the scoring methods, write FAQs. & 50 \\
 \hline
 9 & \textbf{Beta testing.} Organize and conduct tests of the challenge.
 & 25 \\
 \hline
 10 & \textbf{Run the challenge.} Answer participants, attend to the platform (2h/week). & 100 \\
 \hline
11 & \textbf{Prepare the workshops.} Write proposals. Look for tutorial speakers. Select speakers. Create a schedule. Advertise. & 50 \\
 \hline
 12 & \textbf{Competition result analysis.} Compile the results. Produce graphs. Derive conclusions. & 50 \\
 \hline
  13 & \textbf{Reports.} Write reports on the benchmark design, the datasets, and the results of the competition. & 100 \\
 \hline
  14 & \textbf{On-line result dissemination.} Make available on-line the competition result analyses, fact sheets of the competitors's methods, and the workshop slides. & 50 \\
 \hline
  15 & \textbf{Prepare workshop proceedings.} Solicit papers, organize the review process, and edit the papers. & 100 \\
 \hline
  16 & \textbf{Distribute prizes and awards.} & 10 \\
  \hline
\end{tabular}
\caption{Evaluation of man power to organize a challenge (varies from challenge to challenge, should be estimated by the organizing team)}
\label{table:manpower}
\end{center}
\end{table}

\subsection{Resources: sponsors and grant agencies}

As the global cost of competition organisation grows along with the complexity of the data and tasks, proposal and grant writing to find money is essential. By leveraging institutional support and sponsors, organizers will achieve good quality challenges and ensure community participation. More and more universities and national funding agencies\footnote{For instance the University College of London, the National Research Agency in France, the ETH in Switzerland, or the EIT Health in Europe} or scientific societies\footnote{National Science Foundation in the United States, the IEEE Computational Intelligenece Society, or the International Neural Network Society} support competition organization. Building partnership with private companies\footnote{Non-exhaustive list of potential sponsors: Google, Microsoft, Orange, Kaggle, Health discovery corporation} and involving collaborators in scientific consortium is also likely to be very helpful to reduce the financial barriers in organizing challenges.

\section{Conclusion}

In this chapter, we offer some tips and practical details for organizing successful competitions. In the first part, we review the different types of motivations that lead participants to participate in the challenge. Then, we offer advice on how to recruit the scientific community concerned and disseminate the challenge and its results. Finally, we review the different stages of preparation for a challenge, the associated financial and human costs, and the possibilities of fundings.

\adrien{Magali:

- pour completer le tableau 12.1, notamment la partie LAP series (combien de participants en tout) et Challenge AI for industry (combien de participants et dissemination). D’ailleurs si l’étudiante d’Adrien veut compléter ce tableau avec d’autres exemples, elle est la bienvenue)

- pour compléter la liste des sponsors et grant agencies proposée en footnote section 12.3.3 -> si vous avez des idées, vous pouvez les rajouter directement…
}

\bibliographystyle{unsrt}
\bibliography{ref}

\begin{thebibliography}{10}

\bibitem{brandenburger_co-opetition_2011}
Adam~M. Brandenburger and Barry~J. Nalebuff.
\newblock {\em Co-{Opetition}}.
\newblock Crown, July 2011.

\bibitem{abernathy_academic_2001}
Tammy~V. Abernathy and Richard~N. Vineyard.
\newblock Academic {Competitions} in {Science}: {What} {Are} the {Rewards} for {Students}?
\newblock {\em The Clearing House}, 74(5):269--276, 2001.
\newblock Publisher: Taylor \& Francis, Ltd.

\bibitem{pavao2019design}
Adrien Pavao, Diviyan Kalainathan, Lisheng Sun-Hosoya, Kristen Bennett, and Isabelle Guyon.
\newblock Design and analysis of experiments: A challenge approach in teaching.
\newblock In {\em CiML workshop, NeurIPS 2019-33th Annual Conference on Neural Information Processing Systems}, 2019.

\bibitem{bender_challenges_2016}
Eric Bender.
\newblock Challenges: {Crowdsourced} solutions.
\newblock {\em Nature}, 533(7602):S62--S64, May 2016.
\newblock Bandiera\_abtest: a Cg\_type: Nature Research Journals Number: 7602 Primary\_atype: Comments \& Opinion Publisher: Nature Publishing Group Subject\_term: Drug discovery and development Subject\_term\_id: drug-discovery-and-development.

\bibitem{madry_towards_2019}
Aleksander Madry, Aleksandar Makelov, Ludwig Schmidt, Dimitris Tsipras, and Adrian Vladu.
\newblock Towards {Deep} {Learning} {Models} {Resistant} to {Adversarial} {Attacks}.
\newblock {\em arXiv:1706.06083 [cs, stat]}, September 2019.
\newblock arXiv: 1706.06083.

\bibitem{russakovsky_imagenet_2015}
Olga Russakovsky, Jia Deng, Hao Su, Jonathan Krause, Sanjeev Satheesh, Sean Ma, Zhiheng Huang, Andrej Karpathy, Aditya Khosla, Michael Bernstein, Alexander~C. Berg, and Li~Fei-Fei.
\newblock {ImageNet} {Large} {Scale} {Visual} {Recognition} {Challenge}.
\newblock {\em International Journal of Computer Vision}, 115(3):211--252, December 2015.

\bibitem{steijger_assessment_2013}
Tamara Steijger, Josep~F. Abril, Pär~G. Engström, Felix Kokocinski, Tim~J. Hubbard, Roderic Guigó, Jennifer Harrow, and Paul Bertone.
\newblock Assessment of transcript reconstruction methods for {RNA}-seq.
\newblock {\em Nature Methods}, 10(12):1177--1184, December 2013.
\newblock Bandiera\_abtest: a Cg\_type: Nature Research Journals Number: 12 Primary\_atype: Research Publisher: Nature Publishing Group Subject\_term: Genome informatics Subject\_term\_id: genome-informatics.

\bibitem{radivojac_large-scale_2013}
Predrag Radivojac, Wyatt~T. Clark, Tal~Ronnen Oron, Alexandra~M. Schnoes, Tobias Wittkop, Artem Sokolov, Kiley Graim, Christopher Funk, Karin Verspoor, Asa Ben-Hur, Gaurav Pandey, Jeffrey~M. Yunes, Ameet~S. Talwalkar, Susanna Repo, Michael~L. Souza, Damiano Piovesan, Rita Casadio, Zheng Wang, Jianlin Cheng, Hai Fang, Julian Gough, Patrik Koskinen, Petri Törönen, Jussi Nokso-Koivisto, Liisa Holm, Domenico Cozzetto, Daniel W.~A. Buchan, Kevin Bryson, David~T. Jones, Bhakti Limaye, Harshal Inamdar, Avik Datta, Sunitha~K. Manjari, Rajendra Joshi, Meghana Chitale, Daisuke Kihara, Andreas~M. Lisewski, Serkan Erdin, Eric Venner, Olivier Lichtarge, Robert Rentzsch, Haixuan Yang, Alfonso~E. Romero, Prajwal Bhat, Alberto Paccanaro, Tobias Hamp, Rebecca Kaßner, Stefan Seemayer, Esmeralda Vicedo, Christian Schaefer, Dominik Achten, Florian Auer, Ariane Boehm, Tatjana Braun, Maximilian Hecht, Mark Heron, Peter Hönigschmid, Thomas~A. Hopf, Stefanie Kaufmann, Michael Kiening, Denis Krompass, Cedric Landerer, Yannick
  Mahlich, Manfred Roos, Jari Björne, Tapio Salakoski, Andrew Wong, Hagit Shatkay, Fanny Gatzmann, Ingolf Sommer, Mark~N. Wass, Michael J.~E. Sternberg, Nives Škunca, Fran Supek, Matko Bošnjak, Panče Panov, Sašo Džeroski, Tomislav Šmuc, Yiannis A.~I. Kourmpetis, Aalt D.~J. van Dijk, Cajo J. F.~ter Braak, Yuanpeng Zhou, Qingtian Gong, Xinran Dong, Weidong Tian, Marco Falda, Paolo Fontana, Enrico Lavezzo, Barbara Di~Camillo, Stefano Toppo, Liang Lan, Nemanja Djuric, Yuhong Guo, Slobodan Vucetic, Amos Bairoch, Michal Linial, Patricia~C. Babbitt, Steven~E. Brenner, Christine Orengo, Burkhard Rost, Sean~D. Mooney, and Iddo Friedberg.
\newblock A large-scale evaluation of computational protein function prediction.
\newblock {\em Nature Methods}, 10(3):221--227, March 2013.
\newblock Bandiera\_abtest: a Cg\_type: Nature Research Journals Number: 3 Primary\_atype: Research Publisher: Nature Publishing Group Subject\_term: Bioinformatics;Protein function predictions Subject\_term\_id: bioinformatics;protein-function-predictions.

\bibitem{creason_community_2021}
Allison Creason, David Haan, Kristen Dang, Kami~E. Chiotti, Matthew Inkman, Andrew Lamb, Thomas Yu, Yin Hu, Thea~C. Norman, Alex Buchanan, Marijke~J. van Baren, Ryan Spangler, M.~Rick Rollins, Paul~T. Spellman, Dmitri Rozanov, Jin Zhang, Christopher~A. Maher, Cristian Caloian, John~D. Watson, Sebastian Uhrig, Brian~J. Haas, Miten Jain, Mark Akeson, Mehmet~Eren Ahsen, Gustavo Stolovitzky, Justin Guinney, Paul~C. Boutros, Joshua~M. Stuart, Kyle Ellrott, Hongjiu Zhang, Yifan Wang, Yuanfang Guan, Cu~Nguyen, Christopher Sugai, Alokkumar Jha, Jing~Woei Li, and Alexander Dobin.
\newblock A community challenge to evaluate {RNA}-seq, fusion detection, and isoform quantification methods for cancer discovery.
\newblock {\em Cell Systems}, page S2405471221002076, June 2021.

\bibitem{hadaca_consortium_guidelines_2020}
{HADACA consortium}, Clémentine Decamps, Florian Privé, Raphael Bacher, Daniel Jost, Arthur Waguet, Eugene~Andres Houseman, Eugene Lurie, Pavlo Lutsik, Aleksandar Milosavljevic, Michael Scherer, Michael G.~B. Blum, and Magali Richard.
\newblock Guidelines for cell-type heterogeneity quantification based on a comparative analysis of reference-free {DNA} methylation deconvolution software.
\newblock {\em BMC Bioinformatics}, 21(1):16, December 2020.

\bibitem{marbach_wisdom_2012}
Daniel Marbach, James~C. Costello, Robert Küffner, Nicole~M. Vega, Robert~J. Prill, Diogo~M. Camacho, Kyle~R. Allison, {DREAM5 Consortium}, Manolis Kellis, James~J. Collins, and Gustavo Stolovitzky.
\newblock Wisdom of crowds for robust gene network inference.
\newblock {\em Nature Methods}, 9(8):796--804, July 2012.

\bibitem{eicher_challenges_2019}
Tara Eicher, Andrew Patt, Esko Kautto, Raghu Machiraju, Ewy Mathé, and Yan Zhang.
\newblock Challenges in proteogenomics: a comparison of analysis methods with the case study of the {DREAM} proteogenomics sub-challenge.
\newblock {\em BMC bioinformatics}, 20(Suppl 24):669, December 2019.

\bibitem{marot_learning_2021}
Antoine Marot, Benjamin Donnot, Gabriel Dulac-Arnold, Adrian Kelly, Aïdan O'Sullivan, Jan Viebahn, Mariette Awad, Isabelle Guyon, Patrick Panciatici, and Camilo Romero.
\newblock Learning to run a {Power} {Network} {Challenge}: a {Retrospective} {Analysis}.
\newblock {\em arXiv:2103.03104 [cs, eess]}, March 2021.
\newblock arXiv: 2103.03104.

\bibitem{le_artificial_2019}
E.~P.~V. Le, Y.~Wang, Y.~Huang, S.~Hickman, and F.~J. Gilbert.
\newblock Artificial intelligence in breast imaging.
\newblock {\em Clinical Radiology}, 74(5):357--366, May 2019.
\newblock Publisher: Elsevier.

\bibitem{guinney_alternative_2018}
Justin Guinney and Julio Saez-Rodriguez.
\newblock Alternative models for sharing confidential biomedical data.
\newblock {\em Nature Biotechnology}, 36(5):391--392, May 2018.
\newblock Bandiera\_abtest: a Cg\_type: Nature Research Journals Number: 5 Primary\_atype: Correspondence Publisher: Nature Publishing Group Subject\_term: Policy;Research data Subject\_term\_id: policy;research-data.

\bibitem{wilkinson_fair_2016}
Mark~D. Wilkinson, Michel Dumontier, IJsbrand~Jan Aalbersberg, Gabrielle Appleton, Myles Axton, Arie Baak, Niklas Blomberg, Jan-Willem Boiten, Luiz~Bonino da~Silva~Santos, Philip~E. Bourne, Jildau Bouwman, Anthony~J. Brookes, Tim Clark, Mercè Crosas, Ingrid Dillo, Olivier Dumon, Scott Edmunds, Chris~T. Evelo, Richard Finkers, Alejandra Gonzalez-Beltran, Alasdair J.~G. Gray, Paul Groth, Carole Goble, Jeffrey~S. Grethe, Jaap Heringa, Peter A.~C. ’t Hoen, Rob Hooft, Tobias Kuhn, Ruben Kok, Joost Kok, Scott~J. Lusher, Maryann~E. Martone, Albert Mons, Abel~L. Packer, Bengt Persson, Philippe Rocca-Serra, Marco Roos, Rene van Schaik, Susanna-Assunta Sansone, Erik Schultes, Thierry Sengstag, Ted Slater, George Strawn, Morris~A. Swertz, Mark Thompson, Johan van~der Lei, Erik van Mulligen, Jan Velterop, Andra Waagmeester, Peter Wittenburg, Katherine Wolstencroft, Jun Zhao, and Barend Mons.
\newblock The {FAIR} {Guiding} {Principles} for scientific data management and stewardship.
\newblock {\em Scientific Data}, 3(1):160018, March 2016.
\newblock Bandiera\_abtest: a Cg\_type: Nature Research Journals Number: 1 Primary\_atype: Comments \& Opinion Publisher: Nature Publishing Group Subject\_term: Publication characteristics;Research data Subject\_term\_id: publication-characteristics;research-data.

\bibitem{cabili_simplifying_2018}
Moran~N. Cabili, Knox Carey, Stephanie O.~M. Dyke, Anthony~J. Brookes, Marc Fiume, Francis Jeanson, Giselle Kerry, Alex Lash, Heidi Sofia, Dylan Spalding, Anne-Marie Tasse, Susheel Varma, and Ravi Pandya.
\newblock Simplifying research access to genomics and health data with {Library} {Cards}.
\newblock {\em Scientific Data}, 5(1):180039, March 2018.
\newblock Bandiera\_abtest: a Cc\_license\_type: cc\_by Cg\_type: Nature Research Journals Number: 1 Primary\_atype: Comments \& Opinion Publisher: Nature Publishing Group Subject\_term: Medical research;Research data Subject\_term\_id: medical-research;research-data.

\bibitem{nicol_consent_2019}
Dianne Nicol, Lisa Eckstein, Heidi~Beate Bentzen, Pascal Borry, Mike Burgess, Wylie Burke, Don Chalmers, Mildred Cho, Edward Dove, Stephanie Fullerton, Ryuchi Ida, Kazuto Kato, Jane Kaye, Barbara Koenig, Spero Manson, Kimberlyn McGrail, Eric Meslin, Kieran O'Doherty, Barbara Prainsack, Mahsa Shabani, Holly Tabor, Adrian Thorogood, and Jantina~de Vries.
\newblock Consent insufficient for data release.
\newblock {\em Science}, 364(6439):445--446, May 2019.
\newblock Publisher: American Association for the Advancement of Science Section: Letters.

\bibitem{ellrott_reproducible_2019}
Kyle Ellrott, Alex Buchanan, Allison Creason, Michael Mason, Thomas Schaffter, Bruce Hoff, James Eddy, John~M. Chilton, Thomas Yu, Joshua~M. Stuart, Julio Saez-Rodriguez, Gustavo Stolovitzky, Paul~C. Boutros, and Justin Guinney.
\newblock Reproducible biomedical benchmarking in the cloud: lessons from crowd-sourced data challenges.
\newblock {\em Genome Biology}, 20(1):195, September 2019.

\end{thebibliography}

\end{document}